\documentclass{article}

\usepackage[accepted]{icml2019}
\pdfoutput=1

\usepackage{hyperref}

\usepackage{microtype}
\usepackage{graphicx}
\usepackage{subfigure}
\usepackage{booktabs}
\usepackage{caption}

\usepackage{amsmath}
\usepackage{amssymb}
\usepackage{mathtools}
\usepackage{float}

\usepackage{tikz}
\usepackage{tikzscale}
\usetikzlibrary{shapes.geometric,arrows,chains,matrix,positioning,scopes,calc, bayesnet}
\usepackage{pgf}

\usepackage{cleveref}

\usepackage{xcolor}

\usepackage{MarkMathCmds}

\newcommand{\X}{X}
\newcommand{\Y}{Y}
\newcommand{\xo}{\mathbf{x}_1}

\newcommand{\mxp}{\boldsymbol{\mu}_{p_{_1}}}
\newcommand{\sxp}{\mathbf{\Sigma}_{p_{_1}}}
\newcommand{\xt}{\mathbf{x}_t}

\newcommand{\yt}{\mathbf{y}_t}
\newcommand{\at}{\mathbf{A}_t}
\newcommand{\bt}{\mathbf{b}_t}
\newcommand{\s}{\mathbf{S}}
\newcommand{\st}{\s_t}
\newcommand{\ft}{f(\mathbf{x}_t)}

\newcommand{\xtp}{\mathbf{x}_{t+1}}
\newcommand{\bu}{\mathbf{u}}
\newcommand{\muu}{\boldsymbol{\mu}_{\mathbf{u}}}
\newcommand{\su}{\mathbf{\Sigma}_{\mathbf{u}}}
\newcommand{\Z}{Z}

\newcommand{\calcd}[1][]{\mathrm{d}#1}
\newcommand{\Exp}[2]{\mathbb{E}_{#1}\left[#2\right]}
\DeclarePairedDelimiterX{\dbar}[2]{[}{]}{\,#1\,\delimsize\|\,#2\,}
\newcommand{\KL}[2]{\mathrm{KL} \dbar*{#1}{#2}}

\newcommand{\muftp}{\boldsymbol{\mu}_{f_{t+1}}}
\newcommand{\covftp}{\mathbf{\Sigma}_{f_{t+1}}}
\newcommand{\cft}{\mathbf{C}_f(\xt)}

\newcommand{\fmu}{f^{\neq \vu}}

\newcommand{\defas}{\triangleq}

\newcommand{\Dc}{{D_\vc}}
\newcommand{\Dx}{{D_\vx}}

\newcommand{\lb}{\mathcal{L}}

\newcommand{\rft}{\mathbf{f}_t}
\newcommand{\cfut}{\mathbf{C}_{f|\vu}(\xt)}

\newcommand{\minus}{\scalebox{0.75}[1.0]{$-$}}

\usepackage{titlesec}
\titlespacing\section{0pt}{4pt plus 4pt minus 2pt}{-4pt plus 2pt minus 2pt}
\titlespacing\subsection{0pt}{0pt plus 4pt minus 2pt}{-4pt plus 2pt minus 2pt}

\icmltitlerunning{Overcoming Mean-Field Approximations in Recurrent Gaussian Process Models}

\begin{document}

\setlength{\abovedisplayskip}{5pt}
\setlength{\belowdisplayskip}{5pt}

\twocolumn[
\icmltitle{Overcoming Mean-Field Approximations \\ in Recurrent Gaussian Process Models}

\icmlsetsymbol{equal}{*}

\begin{icmlauthorlist}
\icmlauthor{Alessandro Davide Ialongo}{cam,mpi}
\icmlauthor{Mark van der Wilk}{pro}
\icmlauthor{James Hensman}{pro}
\icmlauthor{Carl Edward Rasmussen}{cam}
\end{icmlauthorlist}

\icmlaffiliation{cam}{Computational and Biological Learning Group, University of Cambridge}
\icmlaffiliation{mpi}{Max Planck Institute for Intelligent Systems, T\"ubingen}
\icmlaffiliation{pro}{PROWLER.io}

\icmlcorrespondingauthor{Alessandro Davide Ialongo}{adi24@cam.ac.uk}

\icmlkeywords{Gaussian processes, time series, dynamical systems, variational inference, Gaussian process state-space model, recurrent Gaussian process model}

\vskip 0.3in
]

\printAffiliationsAndNotice{}

\begin{abstract}
We identify a new variational inference scheme for dynamical systems whose transition function is modelled by a Gaussian process. Inference in this setting has either employed computationally intensive MCMC methods, or relied on factorisations of the variational posterior. As we demonstrate in our experiments, the factorisation between latent system states and transition function can lead to a miscalibrated posterior and to learning unnecessarily large noise terms. We eliminate this factorisation by explicitly modelling the dependence between state trajectories and the Gaussian process posterior. Samples of the latent states can then be tractably generated by conditioning on this representation. The method we obtain (VCDT: variationally coupled dynamics and trajectories) gives better predictive performance and more calibrated estimates of the transition function, yet maintains the same time and space complexities as mean-field methods. Code is available at: \href{https://github.com/ialong/GPt}{\nolinkurl{github.com/ialong/GPt}}.
\end{abstract}

\section{Introduction}

Many time series are well explained by assuming their progression is determined by an underlying dynamical system. A model of this dynamical system can be used to make strong predictions of the future behaviour of the time series. We often require predictions of systems with unknown dynamics, be it for economic models of financial markets, or models of physical systems to be controlled with model predictive control or model-based reinforcement learning. Particularly in low-data regimes, estimates of the uncertainty in the predictions are crucial for making robust decisions, as the expected utility of an action can depend strongly on the distribution of the outcome \citep{neumann1944}. A striking example of this is model based policy search \citep{deisenroth2011pilco}.

There are many approaches to time series prediction. Within the machine learning community, autoregressive (AR) \citep{billings2013nonlinear} and state-space (SSMs) models are popular, in part due to the relative ease with which predictions can be made. AR predictions are obtained by a learned mapping from the $H$ last observations to the next one, whereas SSMs model the underlying dynamical system by learning a \emph{transition function} which maps a system \emph{state} forward in time. At each point in time, the state contains sufficient information for predicting both future states and observations. While AR models can be easier to train, SSMs have the potential to be more data efficient, due to their minimal state representation.

We aim to learn complex, possibly stochastic, non-linear dynamical systems from noisy data, following the Bayesian paradigm in order to capture model uncertainty. We use a Gaussian process (GP) prior \citep{rasmussen2006gpml} for the transition function, giving the Gaussian Process State Space Model (GPSSM) \citep{frigola2013gpssm-pmcmc}. Aside from capturing uncertainty, GPs are non-parametric, guaranteeing that our model complexity will not saturate as we observe more data.

Despite the challenge of performing accurate approximate inference in the GPSSM, an impressive amount of progress has been made \citep{frigola2014vgpssm,mchutchon2014nonlinear,eleftheriadis2017gpssm,ialongo2017gpssm,doerr2018probabilistic}, helped along by the development of elegant variational inference techniques \citep{titsias2009,hensman2013} which retain the key non-parametric property of GPs. In this work, we improve variational approximations by critically examining existing methods and their failure modes. We propose a family of non-factorised variational posteriors which alleviate crucial problems with earlier approaches while maintaining the same efficiency. We refer to the proposed approach as VCDT: variationally coupled dynamics and trajectories.

\section{Background}

State-space models are common in machine learning, and appear in many forms. At the most basic level, linear-Gaussian state-space models can be learned by maximum likelihood, combining Kalman smoothing and EM for instance \citep{roweis1999unifying}. Extensions have been developed for deterministic non-linear transitions \citep{ghahramani1999learning}. Recurrent neural networks, like the popular LSTM \citep{hochreiter1996lstm} learn deterministic mappings with state space structure for sequence prediction, which have seen successful recent use in a wide range of tasks including translation \citep{sutskever2014seq}.

\subsection{Bayesian non-parametric approaches}
We are particularly interested in prediction tasks which require good estimates of uncertainty, for example, for use in model-based reinforcement learning systems \citep{deisenroth2011pilco}. These applications distinguish themselves by requiring 1) continual learning, as datasets are incrementally gathered, and 2) uncertainty estimates to ensure that policies learned are robust to the many dynamics that are consistent with a small dataset. Bayesian non-parametric models provide a unified and elegant way of solving both these problems: model complexity scales with the size of the dataset and is controlled by the Bayesian \emph{Occam's razor} \citep{rasmussen2000occam}. In this work, we focus on approximations \citep{titsias2009,hensman2013,matthews2016kl} which offer improved computational scaling with the ability to fully recover the original non-parametric model.

\subsection{Gaussian process State Space Models}
Recurrent models with GP transitions (GPSSM) have been proposed in a variety of ways, each with its own inference method \citep{wang2005gpdm,ko2009gp,turner2010,frigola2013gpssm-pmcmc}. Early methods relied on \emph{maximum a posteriori} inference for either the latent states \citep{wang2005gpdm,ko2009gp} or the transition function \citep{turner2010}. \citet{frigola2013gpssm-pmcmc} presented the first fully Bayesian treatment with a particle MCMC method that sampled over the latent states and the GP transition function. Due to issues with computational complexity and sampling in higher dimensional latent spaces, later attention turned to variational methods.
All variational inference schemes have so far relied on independence assumptions between latent states and transition function \citep{frigola2014vgpssm,mchutchon2014nonlinear,ialongo2017gpssm}, sometimes even factorising the state distribution over time \citep{mattos2016recurrent}. \citet{eleftheriadis2017gpssm} introduced a recognition model to help with optimisation of the variational distribution, while keeping the mean-field assumption. Recently, \citet{doerr2018probabilistic} introduced the first method to account for the dependence between transition function and latent states by using a doubly stochastic inference scheme similar to \citet{salimbeni2017doubly}. However, their approach has severe limitations which we will discuss and show experimentally. \citet{bui2016pepgpssm} investigated Power Expectation Propagation as an alternative approach for fitting factorised approximate posteriors.

\subsection{Variational inference approaches}
Variational methods often make simplifying assumptions about the form of the approximate posterior to improve computational tractability. This may result in significantly biased solutions. The bias is particularly severe if independence assumptions are made where strong correlations are actually present \citep{turner2011two}. Clearly, in dynamical systems, the trajectory of the latent states depends strongly on the dynamics.
Hence, we focus on improving existing variational inference schemes by removing the independence assumption between latent states and transition function. We identify a general method that performs well across varying noise scales.

\section{The Model}

\subsection{State-space model structure}
We model discrete-time sequences of observations $\Y = \left\{\vy_t\right\}_{t=1}^T$, where $\vy_t \in \Reals^E$, by positing that each data point is generated by a corresponding latent variable $\vx_t \in \Reals^D$ (the system's ``state''). These latent variables form a Markov chain, implying that, for any time-step \(t\), we can generate \(\xtp\) by conditioning only on \(\xt\) and the transition function \(f\). We take the system to evolve according to a single, time-invariant transition function, which we would like to infer. 
While our inference scheme provides the freedom to choose arbitrary transition $p(\xtp | f, \xt)$ and observation $p(\vy_t|\vx_t)$ density functions, in keeping with previous literature, we use Gaussians. We also assume a linear mapping between $\vx_t$ and the mean of $\vy_t\given\vx_t$. 
This does not limit the range of systems that can be modelled, though it may require us to choose a higher dimensional latent state \citep{frigola2015bayesian}. We make this choice to reduce the non-identifiabilities between transitions and emissions. To reduce them further we may also impose constraints on the scale of \(f\) or of the linear mapping \(\mathbf{C}\).
This allows the model to be expressed by the following equations: 
\vspace{0.1cm}
\begin{align}
    f &\sim \GP(m(\cdot), k(\cdot, \cdot))  \\
    \vx_1 &\sim \NormDist{\mxp, \sxp}  \\
    \vx_{t+1}\given f, \vx_t &\sim \NormDist{f(\vx_t),  \mathbf{Q}} \\[4pt]
    \vy_t \given \vx_t &\sim \NormDist{\mathbf{C}\vx_t + \mathbf{d}, \mathbf{R}}
\end{align}
where we take the ``process noise'' covariance matrix \(\mathbf{Q}\) to be diagonal, encouraging the transition function to account for all correlations between latent dimensions. 
The crucial difference from a standard state-space model is that we have placed a GP prior over the transition function. Thus, for a given mean function \(m(\cdot)\) and positive-definite covariance function \(k(\cdot, \cdot)\), any finite collection of function evaluations, at arbitrary inputs \(\Z= \left\{\vz_n\right\}_{n=1}^N, \vz_n \in \Reals^D\), will have a Gaussian joint distribution:
\begin{align}\label{eq:GP}
    p(f(\Z)) \defas p(f(\vz_1), \dots, f(\vz_N)) = \NormDist{\boldsymbol{m}_{\Z}, K_{\Z,\Z}} 
\end{align}
where \(\left[\boldsymbol{m}_{\Z}\right]_{n,1} = m(\vz_n)\) and \(\left[K_{\Z, \Z}\right]_{m,n} = k(\vz_m, \vz_n)\).
This gives the joint density of our model:
\begin{equation}
\begin{split}
    p(\X,\Y,f(\X)) &= \\ 
    p(x_1)p(f(&\X))\prod_{t=1}^{T-1}p(\xtp\given f, \xt) \prod_{t=1}^T p(\yt\given\xt) \,.
\end{split}
\end{equation}
In order to draw samples, \(f(\xtp)\) has to be drawn by conditioning on all the previously sampled function values:
\begin{equation}
    p(f(\xtp)\given f(\vx_{1:t})) = \NormDist{f(\xtp)\given \boldsymbol{\mu}_{f_{t+1}}, \mathbf{\Sigma}_{f_{t+1}}} \label{eq:gp-cond}
\end{equation}
\begin{align}
    f(\vx_{1:t}) &\defas \left[f(\xo), \dots, f(\xt)\right]\transpose \label{eq:vecf} \\
    \muftp & \defas m(\xtp)+ K_{\xtp,\vx_{1:t}}K_{\vx_{1:t}, \vx_{1:t}}^{-1} \left(f(\vx_{1:t}) - \boldsymbol{m}_{\vx_{1:t}}\right) \label{eq:mean}\\
    \covftp &\defas K_{\xtp,\xtp} - K_{\xtp,\vx_{1:t}}K_{\vx_{1:t}, \vx_{1:t}}^{-1} K_{\vx_{1:t},\xtp} \,. \label{eq:var}
\end{align}
By multiplying the Gaussian conditionals, we can return to our GP prior as in \cref{eq:GP}. Although the function values are jointly Gaussian, the states \(\X\) are not necessarily so, since the GP's mean and kernel functions specify non-linear relations between inputs \(\xt\) and outputs \(f(\xt)\).

\subsection{Multivariate latent states}
If our state has more than one dimension (i.e. \(D>1\)), we model the system's transitions \(f:\Reals^D \to \Reals^D\) by placing a GP prior on each of the \(f_d:\Reals^D \to \Reals\) dimension-specific functions, and we write:
\begin{equation}
    f(\xt) \defas \left\{f_d(\xt)\right\}_{d=1}^D \,, \hspace{0.2cm}
    p(f(X)) = \prod_{d=1}^D p(f_d(X))
\end{equation}
where each independent GP has its own mean and kernel functions: \(m_d(\cdot), k_d(\cdot, \cdot)\). For ease of notation, we drop subscripts indexing state dimensions \(d\). Since the GP's are independent from each other, each process only has to condition on its own function evaluations. Thus, the multivariate conditional \(p(f(\xtp)\given f(\vx_{1:t}))\) is a diagonal Gaussian where the mean and variance of dimension \(d\) are as in \cref{eq:mean} and \cref{eq:var}, using as mean function \(m_d(\cdot)\) and as kernel function \(k_d(\cdot, \cdot)\).

\subsection{Control inputs}
A control input $\vc_t \in \Reals^{\Dc}$ influences the transition in a Markovian manner. We can therefore view it as being part of an ``augmented'' latent state. This makes our transition function $f: \Reals^{\Dx + \Dc} \to \Reals^\Dx$, and requires only a change to $f$'s kernel and mean functions to be defined over \(\Dx + \Dc\) input dimensions. Our transition probability is now also conditioned on $\vc_t$: $p(\xtp\given f, \xt, \vc_t)$, though, for brevity, we drop $\vc_t$ from the notation, as it is not a random variable and we always condition on it.

\subsection{Computational costs of sampling}
\label{sec:prior-sampling}
Despite the conceptual similarity between the non-parametric GPSSM and parametric non-linear state space models, it is considerably more expensive to sample from the GPSSM prior. The main reason for this can be seen in \cref{eq:gp-cond,eq:vecf,eq:mean,eq:var}: to sample $\vx_{T}$ we have to condition on the $T-1$ previous points. This requires incrementally building up a $T\times T$ matrix inverse, which costs $\BigO\left(T^3\right)$ in time. Since our non-parametric transition function is implicitly defined by the function values \(f(\vx_{1:t})\) sampled so far, it is necessary to condition on all previous samples to ensure our overall function is self-consistent over the whole time series.

This poses a problem for MCMC based inference methods, which rely on drawing samples from distributions closely related to the prior. Prediction from a GP posterior also has cubic cost. One way to avoid this cost is to sample from an adapted GP prior (e.g.~FITC \citep{snelson2005sparse}), which was suggested by \citet{frigola2013gpssm-pmcmc}.

Modifying the model in this way can have unintended consequences \citep{bauer2016understanding}, so we focus on performing inference without assumptions on the structure of the GP or its kernel. Our inference scheme should approximate the correct model, while avoiding the $\BigO\left(T^3\right)$ cost at training time.

\section{Variational inference}

\subsection{General variational bound}
We are interested in approximating the posterior $p(\X, f\given\Y)$\footnote{We abuse notation by directly denoting a density over the infinite-dimensional $f$, even though such a density does not exist. We only ever concern ourselves with finite subsets of \(f\).} using variational inference. The general procedure is to construct a variational lower bound $\lb$ to the log marginal likelihood that has the KL divergence from an approximation to the true posterior as its slack: $\log p(Y) - \lb = \KL{q(\X, f)}{p(\X, f\given\Y)}$. By maximising $\lb$ w.r.t.~$q$, we improve the quality of the approximation.

Practical stochastic variational inference places two requirements on approximating distributions. First, we need to be able to generate samples from them and, second, we need to be able to evaluate their density. 
We start by deriving the lower bound for a general joint approximate posterior $q(\X, f)$, to which we will later add constraints in order to produce different tractability-accuracy trade-offs. Following the usual derivation using Jensen's inequality \citep{blei2017variational}, we obtain
\begin{align}
    &\log p(\Y) \geq \int q(\X, f) \Bigg[\log \prod_{t=1}^T p(\vy_t|\vx_t)+\nonumber \\
    &\qquad \log \frac{p(\vx_1)p(f)\prod_{t=1}^{T-1} p(\vx_{t+1}\given f, \vx_t)}{q(\X, f)} \Bigg]  \calcd f \calcd \X \\
    &= \sum_{t=1}^T \Exp{q(\xt)}{\log p(\vy_t\given\xt)} - \KL{q(f)}{p(f)} +  \nonumber \\
    &\qquad \Exp{q(\X, f)}{\log\frac{p(\vx_1)\prod_{t=1}^{T-1}p(\vx_{t+1}\given f, \xt)}{q(\X\given f)}} \,. \label{eq:elbo-general}
\end{align}

For \(q(f)\) we will use a sparse Gaussian process \citep{titsias2009,hensman2013,matthews2016thesis}. This constraint is ubiquitous in models where the inputs to a GP are random variables \citep{damianou2016variational}, such as the GPLVM \citep{titsias2010bayesian}, or deep GP \citep{damianou2013deep}, as it provides analytical as well as computational tractability. 
A sparse GP approximation specifies a free Gaussian density on the function values $\vu$ at $M$ locations $q(\vu) = \NormDist{\vu; \muu, \su}$, and uses the prior conditional for the rest of the Gaussian process, which we denote $f^{\neq \vu}$. At this time, we let $\X$ depend on the entire process $f$ for generality:
\begin{equation}
    q(f^{\neq \vu}, \vu, \X) = q(\X\given f^{\neq \vu}, \vu) p(f^{\neq \vu}\given \vu) q(\vu) \,.
\end{equation}
Choosing the posterior in this way results in a finite KL between the approximate posterior and the prior, which becomes $\KL{q(\vu)}{p(\vu)}$ \citep{matthews2016kl}.

\subsection{Approximate posterior design choices}
\begin{table*}[t]
\captionsetup{justification=raggedright, singlelinecheck=false, format=hang}
\centering
\begin{tabular}{lccccc}
                             & \(q(\X \given  f)\)  & $\rft$                         & $\st^*$                   &  sampling \\ \hline
1) Factorised - linear          & \(q(\X)\)                   & $\vx_t$                        & $\st$                    &  $\mathcal{O}(T)$ \\
2) Factorised - non-linear      & \(q(\X)\)                   & $K_{\xt,\Z}K_{\Z,\Z}^{-1}\muu$ & $\st + \at\cft\at^\top$ &  $\mathcal{O}(T)$ \\
3) Non-Factorised - non-linear  & \(q(\X\given f)\)                   & $f(\vx_t)$                     & $\st$                    &  $\mathcal{O}(T^3)$ \\
4) VCDT    & \(q(\X\given \vu)\)                   & $K_{\xt,\Z}K_{\Z,\Z}^{-1}\vu$  & $\st + \at\cfut\at^\top$                    &  $\mathcal{O}(T)$
\end{tabular}
\caption{\label{tab:posteriors}Variations of approximate posteriors. \(\at, \bt, \st\) are free parameters in all cases. \(\cft\) is the sparse GP's marginal posterior variance $\cft \defas K_{\xt,\xt} + K_{\xt,\Z}K_{\Z,\Z}^{-1} \left(\su - K_{\Z,\Z}\right) K_{\Z,\Z}^{-1} K_{\Z,\xt}$ whereas \(\cfut\) is the conditional variance of \(\ft\given\vu\): $
\cfut \defas K_{\xt,\xt} - K_{\xt,\Z}K_{\Z,\Z}^{-1} K_{\Z,\xt}$.}
\end{table*}

So far, the only simplifying assumptions have been reducing \(q(\fmu|\vu)\)  to the prior conditional \(p(\fmu|\vu)\) and \(q(\vu)\) to a Gaussian. Now we have to specify the form of \(q(\X|\fmu, \vu)\). Most existing approximate inference schemes focused on factorised posteriors \(q(\X|\fmu, \vu) = q(\X)\) to obtain an efficient sampling scheme or even a closed form for $\lb$. In particular, \citet{frigola2014vgpssm} find the optimal variational \(q(\X)\) by calculus of variations and sample from it
, whereas \citet{mchutchon2014nonlinear} and \citet{ialongo2017gpssm} constrain it to be Markov-Gaussian.
A factorised, Gaussian posterior together with Gaussian emissions leads to a Gaussian optimal $q(\vu)$, which, for certain kernels, gives us a closed-form bound.
\citet{eleftheriadis2017gpssm} reduced the \(\mathcal{O}(TD^2)\) memory cost associated with storing the Markov-Gaussian $q(\X)$ and improved optimisation behaviour by employing a recurrent neural network as a recognition model. This also parallels the work of \citep{krishnan2017structured}, where recurrent neural networks perform smoothing in order to learn a deterministic model of the dynamics.

In our experiments, we show that imposing independence between \(\X\) and \(f\) does in fact encourage over-confident beliefs about the dynamics, which is a well-known phenomenon in variational inference \citep{turner2011two}. To avoid this we focus on approximate posteriors which preserve correlations between $\X$ and $f$ in various forms. We impose one more general constraint: Markovian structure in $\X$. This structure is required to eliminate the impractical amount of correlations that can be expressed between the large number of random variables in $\X$. Furthermore, the true posterior is also Markovian in $\X$:
\begin{equation}
    p(\X|f, \Y) = p(\vx_1 | f, \Y) \prod_{t=2}^{T} p(\xt|f(\vx_{t-1}), \vy_{t:T}) \,.
\end{equation}
Hence the general form of our approximate posterior:
\begin{equation}
    q(\X\given f) = q(\vx_1)\prod_{t=1}^{T-1}q(\vx_{t+1}\given f, \vx_t) \label{eq:approx-posterior-form} 
\end{equation}
where we are free to specify $q(\vx_1)$ and $q(\vx_{t+1}\given f, \xt)$. We can thus simplify the ELBO from \cref{eq:elbo-general} to
\begin{align}
    &\lb = \sum_{t=1}^T \Exp{q(\xt)}{\log p(\vy_t\given\xt)} - \KL{q(\vu)}{p(\vu)} +  \nonumber \\
    &  \minus \KL{q(\vx_1)}{p(\vx_1)} \nonumber \\
    & \minus \sum_{t=1}^{T-1} \Exp{q(f, \xt)}{\KL{q(\xtp | f,\xt)}{p(\xtp | f, \xt)}} \,.
\end{align}
Now we add two restrictions: 1) we use a Gaussian $q(\xtp\given f, \xt) = \NormDist{\xtp; g_t(f, \xt), h_t(f, \xt)}$ to obtain a tractable KL, and 2) we assume $g_t$ and $h_t$ are linear, and depend only on a finite number of points on $f(\cdot)$, $q(\xtp\given f, \xt) = \NormDist{\xtp; \at\rft + \bt, \st^*}$, to make the expectation over $q(f)$ tractable.
We choose $\at$, $\bt$, and $\st$ to be free variational parameters in all cases (see \cref{tab:posteriors} for how $\st$ and $\st^*$ are related). They are crucial to allow information from the observations to make its way into our latent state posterior. 
We now discuss various choices for $\rft$ and $\st^*$, and the influence they have on the flexibility of the approximate posterior, which we summarise in \cref{tab:posteriors}. The densities of all these variational distributions can still be evaluated, and exact samples can be drawn for unbiased evaluation of the bound.

\subsection{A natural choice}
Another way to justify our choice of posterior is that, for Gaussian emissions, the exact posterior ``filtering'' factors: \(p(\xtp | f, \xt, \vy_{1:t+1})\) are Gaussians of precisely the form we propose. In fact, we can compute the corresponding \(\at\), \(\bt\), and \(\st^*\) in closed form:
\begin{align}
\st^* &= (\mathbf{Q}^{-1} + \mathbf{C}\transpose \mathbf{R}^{-1} \mathbf{C})^{-1} \\
\at &= \st^* \mathbf{Q}^{-1} \\
\bt &= \st^* \mathbf{C}\transpose \mathbf{R}^{-1} (\vy_{t+1} - \mathbf{d})\, .
\end{align}
Notice that here \(\at\) and \(\st^*\) are constant through time, reducing the time and memory footprint of the algorithm.
While the exact posterior ``smoothing'' factors \(p(\xtp | f, \xt, \vy_{1:T})\)  are not generally Gaussian or closed-form (hence leaving \(\at,\bt,\st\) free), in settings where the observation noise is small, filtering and smoothing will give very similar results, making the filtering factors very close to optimal. Likewise, if future observations hold little information about the present due to process noise corruption, the filtering factors will provide good approximations to the conditional state posterior.

\subsection{Factorised approximations}
We can recover the existing Gaussian factorised approximation of \citet{mchutchon2014nonlinear} by taking $\rft = \xt$ (see 1 in \cref{tab:posteriors}). The linear relationship between consecutive states implies a linearisation of the transition function. Yet, the true $p(\xtp|f(\xt),\vy_{t+1:T})$ will contain influence from $\xt$ only through the \emph{non-linear} transition function.

We can break the linearisation constraint by correlating $\xtp$ with $\xt$ ``pushed through'' the approximate GP mean (2 in \cref{tab:posteriors}). This results in a non-Gaussian $q(X)$ which is still factorised from $q(f)$. Since this corresponds to marginalising out \(f\) from \(q(\X\given f)\), we let \(\st^* = \st + \at\cft\at^\top\) to allow our state posterior to be more ``uncertain'' precisely where our GP posterior tells us to. If the observations provide sufficient evidence to discount the GP's uncertainty, optimisation will drive \(\at\) and \(\st\) down, giving us a concentrated state posterior.

The main reason we introduce this modified factorised method, is so we can directly isolate the effect of introducing dependence between $\X$ and $f$, compared to simply making $q(\X)$ non-Gaussian.

Both approximate posteriors can be sampled in $\BigO\left(T\right)$ time due to their Markovian structure. The factorised - linear posterior is particularly convenient as it allows us to exploit GPU-optimised banded-matrix operations as in \citet{durrande2019banded}.

\subsection{Direct dependence on $f$}
The factorised, non-linear approximation from the previous section is attempting to indirectly incorporate the learned transition function into the approximation over $\X$. Instead of summarising the information using the mean of $f$, we want to break the factorisation between $\X$ and $f$ by using the actual, stochastic $f$ in the approximation. This is summarised in \cref{tab:posteriors}, line 3. The \emph{actual} function value $f(\vx_t)$ of a function in the posterior is weighed by $\at$ against $\bt$.

Sampling from this approximate posterior poses a challenge. From \cref{eq:approx-posterior-form}, we note that every time a $f(\xt)$ is sampled, it has to be consistent with a single function throughout the entire time series. This means that $f(\xt)$ has to be sampled by conditioning on the function values sampled for earlier time-steps, resulting in an $\BigO\left(T^3\right)$ cost (as per \cref{sec:prior-sampling}). This high cost prohibits learning in long time series. We can circumvent this issue by cutting the approximate posterior into subsequences with lengths $\tau_1, \dots, \tau_n$. The state immediately after a subsequence, e.g.~$\vx_{\tau_1+1}$ is given a non-conditional posterior:
\begin{align}
    q(\X\given f) &= \left[q(\vx_1)\prod_{t=1}^{\tau_1 - 1}q(\vx_{t+1}\given f, \xt)\right] \times \nonumber \\ 
    &\left[q(\vx_{\tau_1+1})\prod_{t=\tau_1+1}^{\tau_1 + \tau_2 - 1}q(\vx_{t+1}\given f, \xt) \right]\dots  \,.
\end{align}
This reduces the cost of sampling to $\BigO\left(\frac{T}{\tau}\tau^3\right)$ (for \(\tau_1, \dots, \tau_n = \tau\)), and allows for mini-batching over subsequences. However, this factorisation catastrophically ``breaks'' long-range dependences, again introducing a potential mismatch between our state and transition function posteriors. In the extreme case of $\tau = 1$, we would obtain an approximate posterior which factorises completely across time, as in \citet{mattos2016recurrent}.

\subsection{VCDT and ``U-dependence''}
The previous section introduces an approximate posterior exhibiting dependence between $\X$ and $f$. This will generally come at a cubic cost since our transition function has potentially as many degrees of freedom as time-steps. Parametric models avoid this cubic cost since they can sample a transition function with fixed degrees of freedom before sampling the entire time series. Conveniently, our sparse variational GP posterior separates the finite degrees of freedom that we can learn from the data, from the rest of the potentially infinite degrees of freedom of our prior.

In order to gain an approximate posterior with a tractable sampling scheme, we choose $q(\X|f) = q(\X|\vu)$ ((4) in \cref{tab:posteriors}). Crucially, we can sample from this posterior in linear-time, as only a single $\vu$ needs to be sampled from $q(\vu)$ for an entire sample trajectory. We are effectively being parametric about our sample trajectories while minimising a KL distance from the full non-parametric model. By the properties of sparse variational GPs \citep{matthews2016kl}, by adding more inducing points we can reduce this distance.

Intuitively, by finely tiling with inducing points the parts of the state-space we are likely to traverse with our sample paths we can fully specify the behaviour of our transition function in those regions. In this case, conditioning on \(\vu\) is sufficient to ``tie'' down the transitions arbitrarily tightly leaving very little to be gained by additionally conditioning on sampled \(f\) values. The elegance of this approach lies in the fact that we can assess how many inducing points are required by finding the saturation point where the converged ELBO values stop improving as more points are added.

\subsection{Comparison to PR-SSM}
\citet{doerr2018probabilistic} were, to the best of our knowledge, the first to consider a non-factorised variational posterior for the GPSSM. Their work, however, has two significant shortcomings. Firstly, \(q(\xtp|f, \xt)\) is taken to be the same as the prior transition. This leaves no free parameters, except for those in \(q(\xo)\) and \(q(\vu)\), to incorporate information from the observations into our posterior: no filtering or smoothing is possible. This gives a good approximation only when the process noise is low and/or the observed sequence is short, as low noise levels can compound in a long sequence. Even in the absence of process noise, using the observations to update our beliefs about the states can help optimisation greatly by effectively setting up easier ``targets'' for the GP to go through. Notice that by setting $\at = \bf{I}$, $\bt = \bf{0}$, and $\st = \mathbf{Q}$ in our posterior transitions we can also force them to match the prior, making PR-SSM a special case of our more flexible parameterisation. 

Secondly, \citet{doerr2018probabilistic} employ a sampling scheme which gives incorrect marginal samples of \(q(\xt)\), even assuming, as they did, that:
\begin{equation}
    p(f(\vx_1), \dots,f(\vx_{T-1})|\vu) = \prod_{t=1}^{T-1} p(\ft|\bu) \,. \label{eq:fitc_assum}
\end{equation}
Despite this conditional factorisation assumption, once \(\vu\) is integrated out, as in PR-SSM, the \(f(\xt)\)'s become correlated. Thus, in order to correctly sample \(f(\xt)\), we need to condition on all previous samples \(f(\vx_{1:t-1})\), introducing a cubic time cost. This is ignored in PR-SSM and samples from the posterior over \(f\) are effectively drawn \emph{independently} for every time-step \(t\). The resulting samples from \(q(\xt)\) are biased. A mismatch is thus introduced between the functional form of the approximate posterior (which has correlations between function values), and the samples we use to compute its expectations. Hence, PR-SSM's objective does not correspond to a valid variational lower bound.

We solve this by explicitly sampling \(\vu\) and conditioning on it. Conditioning on the same \(\vu\) enforces consistency of the dynamics along a sample trajectory. However, we also wish to avoid the factorisation assumption of \cref{eq:fitc_assum} as that does not generally correspond to a valid GP prior. Explicitly assuming \(q(\X\given f) = q(\X\given\vu)\) we obtain a valid variational bound where we only need to condition on \(\vu\) to sample \(\X\), giving us an \(\BigO\left(T\right)\) cost while maintaining a full GP prior.

\section{Experiments}

In the experiments we set out to test how the proposed posterior (VCDT) compares with the Factorised - linear, Factorised - non-linear, and PR-SSM \citep{doerr2018probabilistic} approximations. 
Each model serves as a ``control'' for a specific experimental question:
\begin{enumerate}
    \item Factorised - linear: what is gained by making our approximation non-linear? I.e. if our posterior \(q(\X)\) were not jointly Gaussian, but only its Markovian conditional factors.
    \item Factorised - non-linear: what is gained by introducing dependence between our states and our transition function? I.e. if we do not marginalise both $f(\xt)$ and $\vu$ from our posterior over $\X$.
    \item PR-SSM: what is gained, or lost, by forcing our model to find transition functions that explain the data well, without accounting for process noise?
\end{enumerate}
To answer these questions, we test model calibration and predictive performance in a number of environments: the ``kink'' function, a set of five system identification benchmark datasets (also used in \citep{doerr2018probabilistic}), and a cart and pole system.

\subsection{``Kink'' Function}
\begin{figure}[htpb]
\centering
\includegraphics[width=0.3\textwidth]{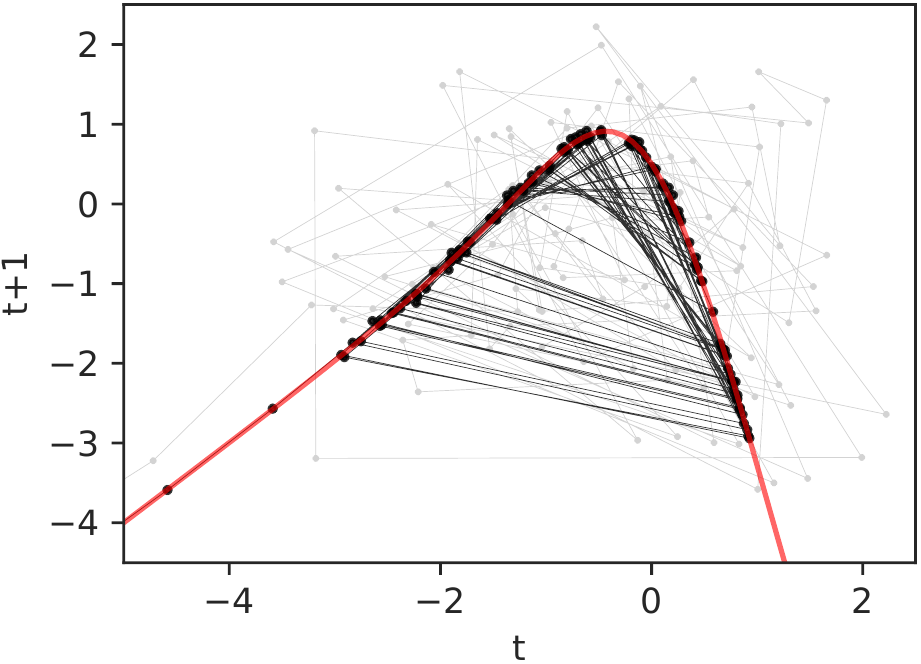}
\caption{``Kink'' data. In red is the true system transition function, in black are the latent states and in grey the observations. Each point's $(x,y)$ coordinates are a pair $(\xt, \xtp)$ or $(\yt, \vy_{t+1})$ for some index $t$. Such pairs are connected through time, highlighting trajectories through the state-space. A sequence of 120 latent states were sampled beginning from a standard normal $\xo$. Their trajectory was perturbed by Gaussian noise with s.d. $0.05$. Observations were obtained by adding Gaussian noise with s.d. $\sqrt{0.8}$ to the latents.}
\label{fig:kink_data}
\vspace{-0.3cm}
\end{figure}

We generate data according to the transition function displayed in red in \cref{fig:kink_data}: \[f(x) = 0.8 + (x + 0.2)(1 - \frac{5}{1 + \exp{(-2x)}})\,.\] This function generates cyclic trajectories through the state-space, as it combines two linear segments with slope greater than \(1\) (left) and less than \(-1\) (right). It also contains a non-linearity at the joining of these segments, giving it a variety of dynamics and serving as an interesting testing environment. We train our models on the data described in \cref{fig:kink_data}, fixing their emissions ($\mathbf{C}, \mathbf{d}, \mathbf{R}$) to the true, generative ones so as to enable direct latent space comparison\footnote{This is to side-step any scale invariances built into our model.}. 

\begin{figure}[htpb]
\centering
\includegraphics[width=0.3\textwidth]{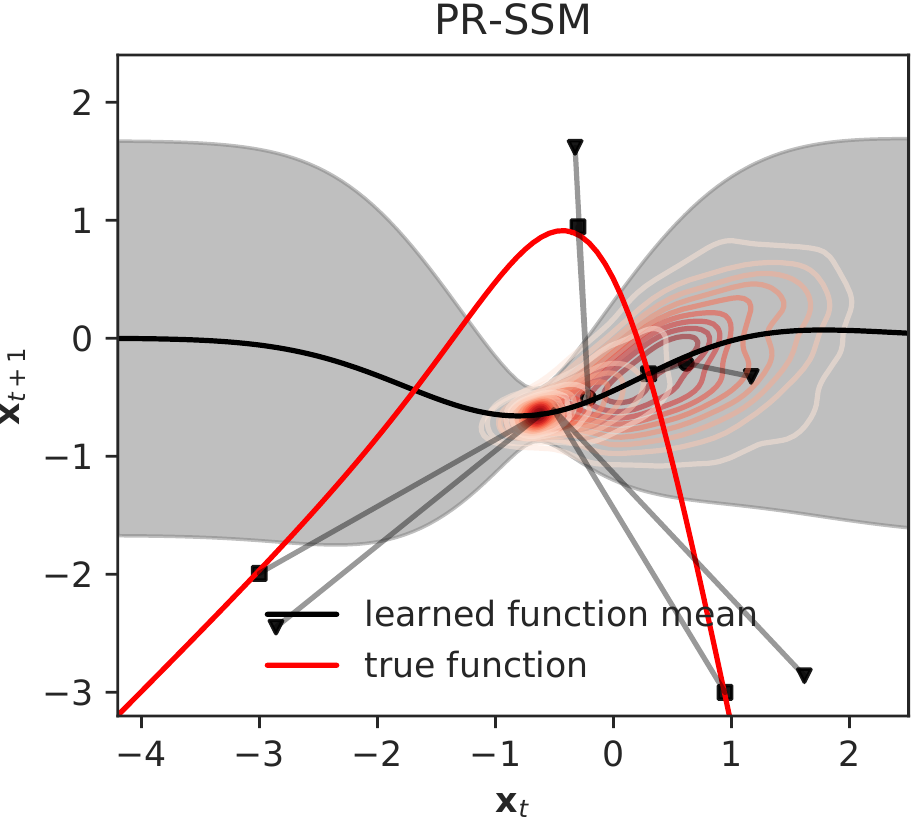}
\caption{PR-SSM fails to learn in the presence of process noise, for moderately long time series ($T=120$). Plot elements as in \cref{fig:kink_trip}.}
\label{fig:prssm_kink}
\vspace{-0.3cm}
\end{figure}

\begin{figure*}
\centering
\includegraphics[width=0.95\textwidth]{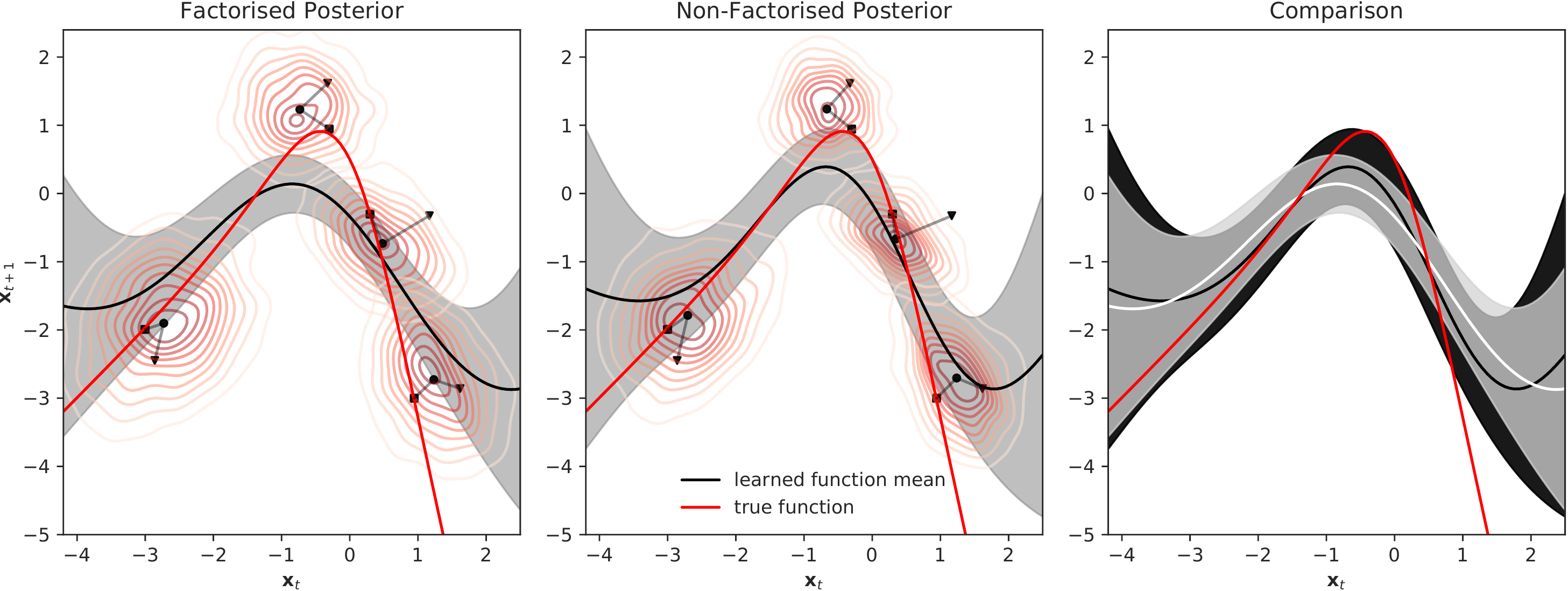} 
\vspace{-0.1cm}
\caption{Posterior miscalibration due to factorisation. \textbf{Left}: fit for the factorised \(q(\X)q(f)\); \textbf{Middle}: fit for VCDT \(q(\X\given \vu)q(f)\); \textbf{Right}: superimposed posteriors (grey is factorised, black is VCDT). \textbf{All subplots}: the shaded regions indicate a \(3\sigma\) confidence interval and the contour plots (left and middle) show the pairwise posteriors \(q(\xt, \xtp)\) over some test latent states. The \((x,y)\) coordinates of the squares, triangles and circles correspond to, respectively: the true latent states \((\xt,\xtp)\), the observed states \((\yt,\vy_{t+1})\) and the means of the marginal posteriors \((\Exp{q(\xt)}{\xt},\Exp{q(\xtp)}{\xtp})\).}
\label{fig:kink_trip}
\end{figure*}

In order to make inference challenging, we add a small amount of process noise (s.d. $0.05$) and a significant amount of observation noise (s.d. $\sqrt{0.8}$). As we can see from \cref{fig:prssm_kink}, in this regime PR-SSM is unable to learn, given that its posterior encourages it to find trajectories with no process noise. Indeed, regardless of which data we are training on, during optimisation, PR-SSM always drives the process noise to zero. This leads it to converge to poor solutions when process noise exists and/or to drive up the observation noise (across all experiments, PR-SSM learned the highest values for the observation noise). Even initialising PR-SSM at the solution found by other methods leads to the same sub-optimal fit of \cref{fig:prssm_kink}. On the other hand, the GPSSM models can handle process noise and recover inflated estimates of it in order to ascribe some transitions to noise and justify smoother dynamics. The Factorised - linear model's estimates of the process noise are consistently the highest (across many settings of noise parameters and sequence lengths), followed by the Factorised - non-linear model. As can be seen in  \cref{fig:kink_trip}, VCDT finds the most well-calibrated posterior. The factorised approaches (the linear one is shown, though they gave a very similar fit) result in a posterior that is not only less accurate, but also more confidently wrong. In general, the posteriors found by the factorised approaches favour higher process noise to achieve smoother, over-confident dynamics. This is consistent with the behaviour discussed in \citep{turner2011two}: mean-field variational inference favours more concentrated posteriors. By matching our true posterior's structure more closely and modelling the dependence between \(\X\) and $f$, we can overcome this issue.

\subsection{System identification benchmarks}
We also train our models on five benchmark datasets taken from \citep{demoor1997daisy}. Test set results are reported in \cref{tab:si_res}. As in \citep{doerr2018probabilistic}, we train on the first half of the sequence, we normalise the data, and we use a 4-dimensional latent state (plus 1 non-stochastic dimension for the control inputs). No mini-batching was used, the bound is evaluated using 100 samples from the posterior, and we use 100 variational inducing points. The test results are for predictions 30 steps into the future from the end of the training sequence.

We initialise all models at the solution found by the Factorised - non-linear method to assess differences in optima of the variational objectives.
This turned out to be necessary for PR-SSM as learning on such long sequences, without mini-batching, proved very challenging without any filtering from the observations. As can be seen from \cref{tab:si_res}, PR-SSM can often find good solutions despite its constraints. This is particularly true for low noise regimes, as in the ``Drive'' dataset. The GPSSM models are more robust in general though, with VCDT in the lead among GPSSMs for NLPP in all but the last two datasets. No error-bars are displayed since the initialisation was deterministic, no mini-batching was used, and 100 samples from the posterior gave low variance estimates of the training objective. To compute the test statistics accurately, we used many more samples ($10^5$).

\begin{table*}
\label{sample-table}
\vspace{0.05cm}
\begin{center}
\begin{small}
\begin{sc}
\begin{tabular}{lccccr}
\toprule
Model & Actuator & Ballbeam & Drive & Dryer & Gas Furnace \\
\midrule
Factorised - linear           & -0.364; 0.154 & -0.486; 0.075 & 0.770; 0.439 & -0.709; 0.098 & -0.296; 0.170 \\
Factorised - non-linear    & -0.641; 0.142 & -1.379; 0.073 & 0.283; 0.246 & -1.310; 0.049 & 0.264; 0.168 \\
VCDT                              & -0.644; 0.141 & -1.395; 0.072 & 0.238; 0.285 & -1.282; 0.050 & 0.377; 0.169 \\
PR-SSM (initialised)       & 3.653; 1.976 & 24.765; 1.118 & -0.649; 0.139 & -1.265; 0.053 & 0.144; 0.162 \\
\bottomrule
\end{tabular}
\end{sc}
\end{small}
\end{center}
\vspace{-0.2cm}
\caption{Test set performance (NLPP; RMSE) for the system identification datasets. Lower is better.}
\vspace{-0.3cm}
\label{tab:si_res}
\end{table*}

\subsection{Cart and pole}
\begin{figure}[h]
\centering
\includegraphics[width=0.45\textwidth]{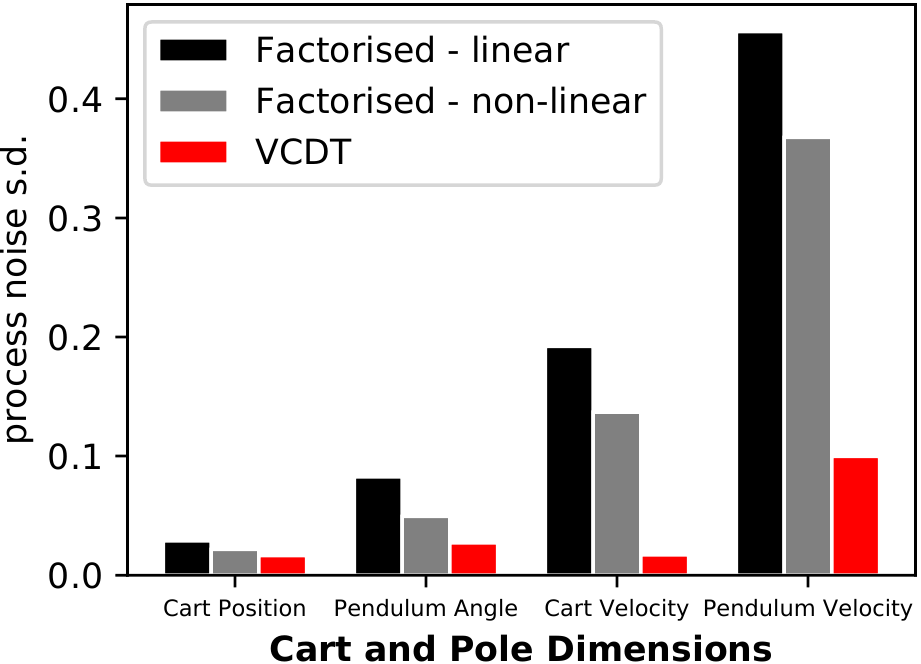}
\caption{Lower values of the learned process noise standard deviation.}
\vspace{-0.18cm}
\label{fig:cp_noise}
\end{figure}

Finally, we consider a cart and pole system (or inverted pendulum). We add Gaussian noise to the states in order to obtain our observations. Standard deviations are: 0.03; 0.1047; 0.4; 1.3963; for the respective dimensions of our system: cart position, pendulum angle, cart velocity and angular velocity of the pendulum. The noise s.d. levels correspond to 3cm; 6 degrees; 3cm/0.075sec; 6 degrees/0.075sec. Results for models fit using 100 samples from the posterior (per optimisation step) and 300 inducing points are shown in \cref{tab:cp_res}.
Because there is no process noise and the training sequences are short (20 time-steps), we see that PR-SSM performs the best in terms of predictive performance. Of course PR-SSM can be viewed as a special case of our approximation, albeit with a mismatched sampling scheme. Running the experiment with the correct sampling scheme and fixing the variational parameters to: $\at = \bf{I}$, $\bt = \bf{0}$, and $\st = \mathbf{Q}$, we learn a solution with the same performance as PR-SSM.
In the GPSSM models, VCDT performs the best, and, as can be seen in \cref{fig:cp_noise}, it also recovers the lowest noise levels, using the dynamics to explain more of the structure in the data.

\begin{table}[htpb]
\begin{center}
\begin{small}
\begin{sc}
\begin{tabular}{lcccr}
\toprule
Model & NLPP & RMSE \\
\midrule
Factorised - linear                    & 2.268 & 3.548 \\
Factorised - non-linear             & 1.847 & 2.974 \\
VCDT                                       & 0.694 & 2.139 \\
PR-SSM                                   & -0.049 & 1.613 \\
PR-SSM (corrected sampling) & -0.053 & 1.628 \\
\bottomrule
\end{tabular}
\end{sc}
\end{small}
\end{center}
\caption{Test set performance for the Cart and Pole dataset. Results averaged over 30 different test sequences (prediction 20 steps into the future). Lower is better.}
\vspace{-0.2cm}
\label{tab:cp_res}
\end{table}

\section{Conclusion}

The GPSSM is a powerful formalism. The main challenge is to perform fast, accurate inference. Variational inference maintains many of the benefits of using non-parametric models, but faces particular difficulties in time series models due to their sensitivity to factorisation assumptions. Naive non-factorising approximations regain the poor computational scaling that variational methods were introduced to avoid. By exploiting the low-rank structure of the approximate GP posterior, we were able to construct a non-factorised posterior with the desired computational scaling. This often leads to better predictive performance, calibration, and an improved estimation of model parameters.   

\newpage

\section*{Acknowledgements}

ADI would like to acknowledge the generous support of the Cambridge-T\"ubingen and Qualcomm Innovation fellowships.

\bibliography{references}
\bibliographystyle{icml2019}

\end{document}